\documentclass[letterpaper, 10 pt, conference]{ieeeconf}  

\IEEEoverridecommandlockouts                              

\usepackage[acronym,nonumberlist]{glossaries}
\usepackage{amssymb}
\usepackage{tabularx}
\usepackage{subcaption}

\providecommand{\GlsX}[1]{\Glsentrydesc{#1} (\glsentrytext{#1})}

\newacronym[shortplural=GMMs]{GMM}{GMM}{Gaussian mixture model}
\newacronym[shortplural=HMMs]{HMM}{HMM}{hidden Markov model}
\newacronym[shortplural=DNNs]{DNN}{DNN}{deep neural network}
\newacronym[shortplural=SVDs]{SVD}{SVD}{singular value decomposition}
\newacronym{RNN}{RNN}{recurrent neural network}
\newacronym{GAN}{GAN}{generative adversarial network}
\newacronym{RL}{RL}{reinforcement learning}
\newacronym{AI}{AI}{artificial intelligence}
\newacronym[shortplural=DTs]{DT}{DT}{decision tree}
\newacronym{SMT}{SMT}{satisfiability modulo theories}
\newacronym{IL}{IL}{Imitation Learning}
\newacronym[shortplural=CNNs]{CNN}{CNN}{convolutional neural network}
\newacronym[shortplural=DQNs]{DQN}{DQN}{deep Q-network}
\newacronym[shortplural=MDPs]{MDP}{MDP}{Markov decision process}
\newacronym[shortplural=POMDPs]{POMDP}{POMDP}{partially observable Markov decision process}
\newacronym[shortplural=CMDPs,longplural={constrained Markov decision processes}]{CMDP}{CMDP}{constrained Markov decision process}
\newacronym{IPPO}{IPPO}{independent proximal policy optimization}
\newacronym{PPO}{PPO}{proximal policy optimization}
\newacronym{MARL}{MARL}{multi-agent reinforcement learning}
\newacronym{CTDE}{CTDE}{centralized training with decentralized execution}
\newacronym{DecPOMDP}{Dec-POMDP}{decentralized partial-observable Markov  decision  processes}
\newacronym{UAV}{UAV}{unmanned aerial vehicle}
\newacronym{MAPPO}{MAPPO}{multi-agent proximal policy optimization}
\newacronym{MADDPG}{MADDPG}{multi-agent deep deterministic policy gradient}
\newacronym{MAAC}{MAAC}{multi-actor-attention-critic}
\newacronym{IQL}{IQL}{independent Q-learning}
\newacronym{IA2C}{IA2C}{independent advantage actor critic}
\newacronym{SQDDPG}{SQDDPG}{Shapley Q-value deep deterministic policy gradient}
\usepackage[usenames,dvipsnames]{xcolor}
\usepackage{pifont}
\usepackage{algorithm}
\usepackage{algpseudocode, amssymb}
\usepackage{booktabs}
\usepackage{multicol,ragged2e}
\usepackage{graphicx}
\usepackage{lineno,hyperref}

\hypersetup{%
linktocpage=true, 
colorlinks=false,
pdfborder={0 0 0}, 
breaklinks=true, pdfpagemode=UseNone, pageanchor=true, pdfpagemode=UseOutlines,%
plainpages=false, bookmarksnumbered, bookmarksopen=true, bookmarksopenlevel=1,%
hypertexnames=true, pdfhighlight=/O,
pdftitle={An Introduction to Multi-Agent Reinforcement Learning and Review of its Application to Autonomous Mobility},%
pdfauthor={Lukas M. Schmidt and Johanna Brosig},%
pdfsubject={SafeDQN},%
pdfkeywords={},%
pdfcreator={pdfLaTeX},%
pdfproducer={LaTeX with hyperref}%
}

\usepackage{amsmath}
\usepackage{import}
\usepackage{varioref}
\usepackage{cite}

\definecolor{betterred}{rgb}{0.8, 0.15, 0.15}
\newcommand{\red} [1]{#1}
\newcommand{\blue} [1]{#1}
\definecolor{britishracinggreen}{rgb}{0.0, 0.5, 0.15}

\definecolor{GoldenRod}{rgb}{0.85, 0.65, 0.13}
\definecolor{treecyan}{rgb}{0.262, 0.8, 0.6}

\title{\LARGE \bf
An Introduction to Multi-Agent Reinforcement Learning \\ and Review of its Application to Autonomous Mobility
}

\author{Lukas M. Schmidt$^{1*}$, Johanna Brosig$^{1*}$, Axel Plinge$^{1\dagger}$, 
Bjoern M. Eskofier$^{2}$, 
and Christopher Mutschler$^{1}$
\thanks{$^{1}$~Fraunhofer IIS, Fraunhofer Institute for Integrated Circuits IIS, Nuremberg, Germany. \texttt{\{firstname\}.\{lastname\}@iis.fraunhofer.de}}
\thanks{$^{2}$~Friedrich-Alexander-Universität Erlangen-Nürnberg (FAU), Erlangen, Germany. \texttt{bjoern.eskofier@fau.de}}
\thanks{* Equal contribution. $\qquad$ $\dagger$ Corresponding Author.}
}%
\begin{document}

\maketitle
\thispagestyle{empty}
\pagestyle{empty}

\begin{abstract}
Many scenarios in mobility and traffic involve multiple different agents that need to cooperate to find a joint solution. Recent advances in behavioral planning use Reinforcement Learning to find effective and performant behavior strategies. However, as autonomous vehicles and vehicle-to-X communications become more mature, solutions that only utilize single, independent agents leave potential performance gains on the road. Multi-Agent Reinforcement Learning (MARL) is a research field aiming to find optimal solutions for multiple agents interacting with each other. This work \red{gives} an overview of the field to researchers in autonomous mobility. We first explain MARL and introduce important concepts. Then, we discuss the central paradigms that underlie MARL algorithms and give an overview of state-of-the-art methods and ideas in each paradigm. With this background, we survey applications of MARL in autonomous mobility scenarios and give an overview of existing scenarios and implementations.
\end{abstract}

\glsresetall

\section{Introduction}
\label{sec:introduction}

\begin{figure}[t]%
    \centering%
    \includegraphics[width=\linewidth]{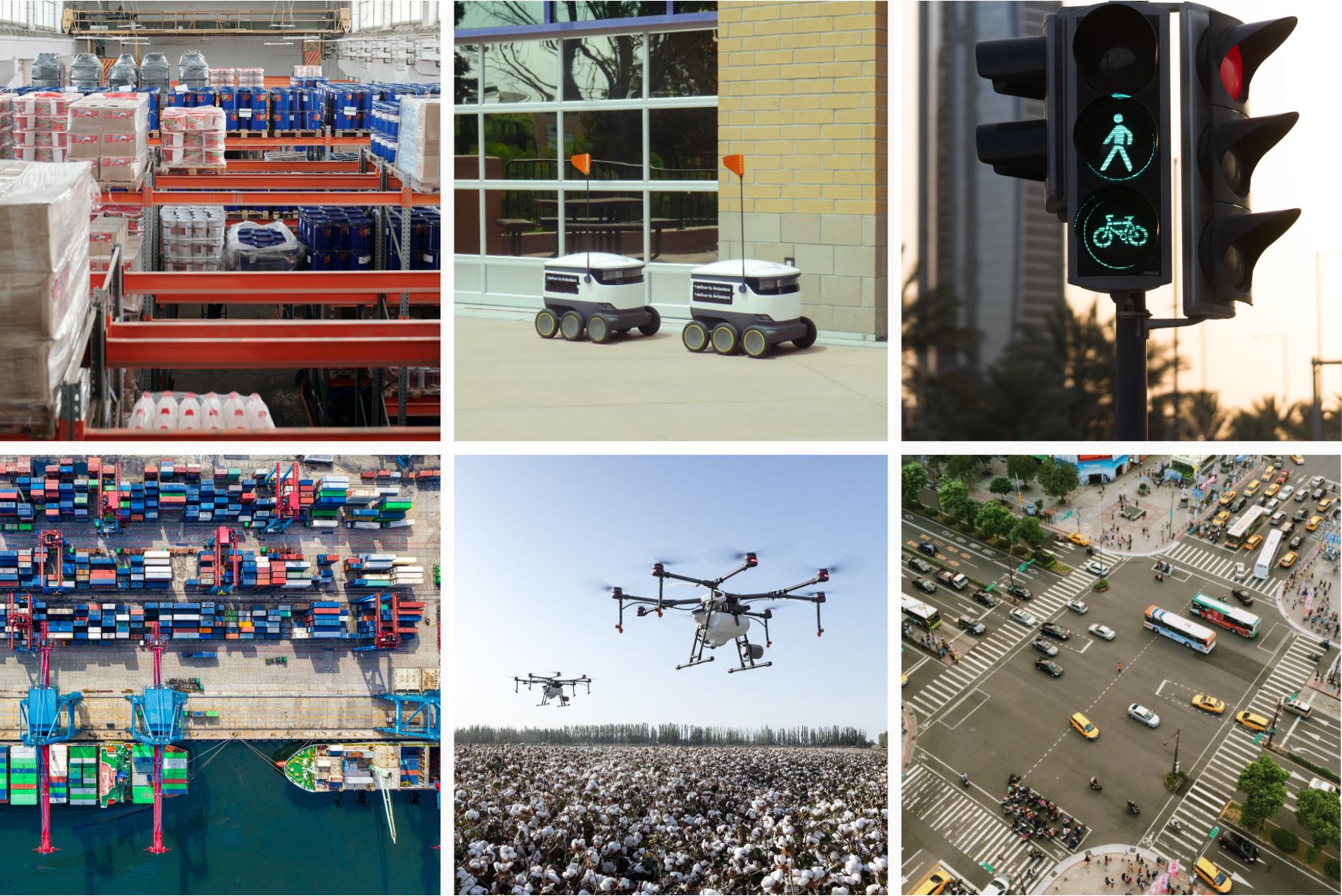}%
    \caption{\Gls{MARL} applications: Intelligent warehouses and logistics, drone fleets for delivery and agriculture, traffic flow control, and automated driving (image source: Pexels/Pixabay).}%
    \label{fig:apps}%
\end{figure}

\Gls{AI} is increasingly being deployed in many applications and devices operating in a connected and cooperative environment. This is particularly relevant in autonomous mobility and traffic scenarios, like the ones shown in Fig.~\ref{fig:apps}, where multiple agents need to interact to find a common solution. As cooperation and communication between the agents are important in these problems, single-agent solutions, such as \gls{RL}, often fall short of expectations. Instead, \gls{MARL} deals with multi-agent problems and aims to find policies that help multiple vehicles reach their individual and common objectives.


\Gls{MARL} aims at finding optimal strategies for agents in settings where multiple agents interact in the same environment. This allows a variety of new solutions that build on concepts such as cooperation~\cite{Oroo19} or competition~\cite{mordatch2017emergence}. However, multi-agent settings also introduce challenges, including potentially missing or inadequate communication, difficulties in credit assignment to agents, and environment non-stationarity. This survey highlights key principles and paradigms in \gls{MARL} research and gives an overview of possible applications and challenges for \gls{MARL} in autonomous mobility.

The paper is structured as follows. Basics of \gls{RL} are provided in Sec.~\ref{sec:basics} before central concepts of \gls{MARL} in Sec.~\ref{sec:marl-basics}. Sec.~\ref{sec:marl-algos} summarizes state-of-the-art \gls{MARL} algorithms and explains core paradigms. Sec.~\ref{sec:applications} gives an overview of  application fields. 
Sec.~\ref{sec:conclusion} concludes.

\section{Basics: Reinforcement Learning} 
\label{sec:basics}

Single-agent RL is designed to solve a class of problems known as Markov decision processes (MDPs).
In an MDP, a single agent interacts with an environment by taking certain actions $a$ depending on the state $s$ of the environment. These actions may change the state of the environment. The agent receives a reward $r$ and aims at maximizing the expected return, i.e., the sum of rewards gained in the environment discounted by a factor $\gamma$ (that weights future rewards) by finding an optimal policy $\pi^*$ (i.e., the \textit{behavior}; ${}^*$ denotes optimality) that maps states to actions. A single step in the environment is known as a transition sample ($s$, $a$, $r$, $s'$)~\cite{sutton1998reinforcement}.

We describe an MDP by a quin-tuple $(S,A,P,R, \gamma)$ where $S$ and $A$ refer to state and action space, $P$ denotes the transition probability from a state $s$ to a successor state $s'$ given an action $a$, $R$ denotes the reward function received by the agent for the transition from $(s,a)$ to $s'$, and $\gamma \in [0;1]$ is the discount factor that trades instantaneous over future rewards.

Two major paradigms in \gls{RL} are value-based and policy gradient methods. Value-based methods, such as DQN~\cite{mnih2015human}, estimate the utility of an action using a parametrized Q-network $Q_\theta$; the policy is then formed by greedily following the actions with the highest estimated values. Instead, policy gradient methods directly optimize a parametrized policy $\pi_\theta$ to take actions that lead to a high reward more likely. Recent work in \gls{RL}, e.g., PPO~\cite{schulman2017proximal} combines both paradigms into an actor-critic architecture that uses a value-based critic to improve updates to the actor (or policy).



\section{Multi-Agent Reinforcement Learning}
\label{sec:marl-basics}
In \gls{MARL}, multiple agents are concurrently optimized to find optimal policies.
Compared to the single-agent \gls{RL} settings, this introduces several important differences and possibilities, which we want to highlight in this section.

\subsubsection{Observability in MARL}
In \gls{RL}, observability describes whether an agent can perceive the entire environment state fully or partially. A classical MDP is defined to be fully observable: The agent directly perceives the environment state $s$. Partial observability is described by a \glspl{POMDP} and modeled by adding a (potentially lossy) observation function converting an environment state $s$ to an observation $o$.
\glspl{POMDP} often present more challenging, real-world applications. For applied research, some scenarios (i.e., robots inside a controlled warehouse environment with enough sensors) can be assumed to be fully observable. Most real-world mobility applications, like traffic scenarios, are partially observable.

\Gls{MARL} provides the opportunity to combine partial observations from multiple agents to gather more information. Several paradigms of \gls{MARL} research, like communication or cooperation, explicitly or implicitly use this opportunity to improve performance. Many MARL methods use Recurrent Neural Networks (RNNs) to aggregate observations over time into more complete observations~\cite{Foerster2016}.

\subsubsection{Centrality}
Different MARL settings can be distinguished by how much agents can communicate with each other. In a fully decentralized problem, agents have no means of communication, while a fully centralized setting allows one central entity to perceive and control all agents.
This distinction affects whether the state and action spaces for all agents are disjoint (decentralized) or joint (fully centralized).
Most MARL algorithms are neither decentralized nor fully centralized; we explain different communication and interaction paradigms in Sec.~\ref{sec:marl-algos}.

\subsubsection{Heterogeneous vs. Homogeneous agents}

In general, \gls{MARL} does not require all agents to be similar. Heterogeneous agents can have different observation and action spaces, like cars and traffic lights.
However, in many \gls{MARL} environments, the agents are homogeneous, meaning they have the same state and action space. For example, many cars require similar behavior in an autonomous driving setting. For homogeneous agents, parameter sharing can be introduced to effectively jointly learn their policies.
This is advantageous as the amount of trainable parameters is reduced, allowing for shorter training times. More efficient training is enabled as the experiences of all agents are used. However, different behavior at execution time is still possible because each agent has unique observations.


\subsubsection{Cooperative vs. Competitive}
An important difference between MARL environments is how the goals of each agent relate to each other. This can be divided into fully cooperative, fully competitive, and mixed cooperative-competitive.

In cooperative settings, all agents aim to achieve a common goal. For example, all agents want to reach their destination undamaged. 
Usually, all  agents  share  a  common  reward  function \cite{Zhang2021}.
This work focuses on cooperative \gls{MARL} because autonomous mobility scenarios are best characterized by shared goals (e.g., reaching each agent's destination unharmed) rather than by competitive objectives.

By contrast, competitive settings, e.g., card or board games, are usually modeled as zero-sum Markov games where the return of all agents sums to zero \cite{Canese2021}. Mixed settings combine cooperative and competitive agents that use general-sum returns. For instance, in team games, agents have to cooperate with their teammates while competing with the opposing teams~\cite{Canese2021}.
Littmann et al.~\cite{Litt2001} state that optimal play can be guaranteed against an arbitrary opponent in competitive settings, whereas strong assumptions must be made to guarantee convergence for cooperative environments.

\subsubsection{Scalability}
While simple \gls{MARL} problems have a relatively small number of agents, many real-world mobility scenarios involve hundreds or thousands of individual traffic participants. Modern simulators, like SUMO~\cite{SUMO2018} or CityFlow~\cite{ZhangFLDZZ00JL19}, can simulate entire neighborhoods or small cities and model how hundreds of traffic lights affect the traffic flow.
Consequently, algorithms must scale to these large agent numbers correctly and efficiently. For example, a simple centralized entity that uses the joint observation and action space of all agents would be impossible to train. In contrast, independent agents (although less performant for a smaller number of agents) could still operate. This problem is known as scalability, and we discuss potential solutions in Sec.~\ref{sec:marl-algos}.

\begin{figure*}
\begin{subfigure}[t]{0.24\textwidth}
\centering
\caption{Decentralized}
\includegraphics[width=.86\linewidth]{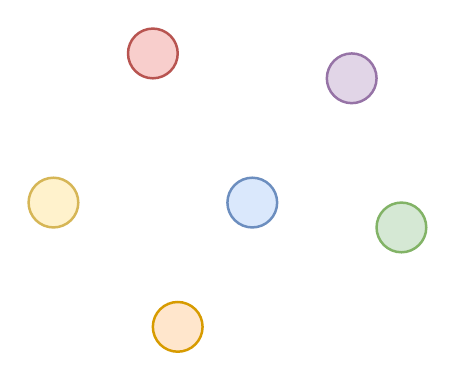}
\end{subfigure}
\hfill
\begin{subfigure}[t]{0.24\textwidth}
\centering
\caption{Credit assignment}
\includegraphics[width=.86\linewidth]{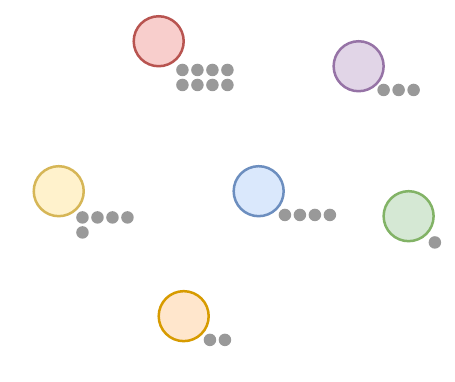}
\end{subfigure}
\hfill
\begin{subfigure}[t]{0.24\textwidth}
\centering
\caption{Communication}
\includegraphics[width=.92\linewidth]{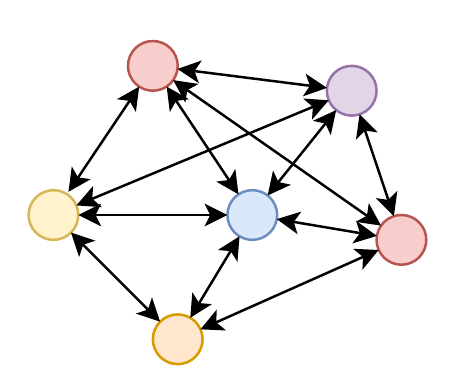}
\end{subfigure}
\hfill
\begin{subfigure}[t]{0.24\textwidth}
\centering
\caption{{CTDE}}
\includegraphics[width=.92\linewidth]{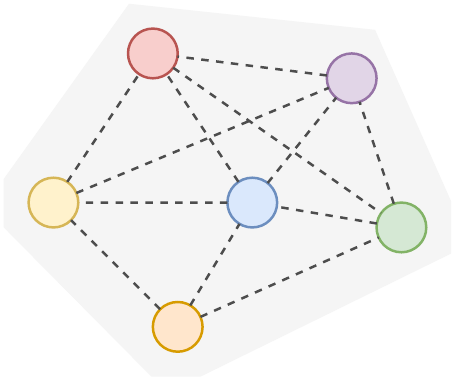}
\end{subfigure}
\vspace{.5ex}
\centering
\caption{Four categories of \gls{MARL} approaches based on communication and cooperation: (a) In a decentralized setting, agents (colored circles) operate and learn independently; (b) Credit assignment techniques use an explicit mechanism that assigns the reward (grey circles) to individual agents; (c) Agents operate independently but have access to a communication medium to exchange messages (black arrows). (d) In \GlsX{CTDE}, agents use a shared critic (gray background) to improve performance during training. After training, the critic is no longer used, and agents operate independently.
}
\label{fig:paradigms}    
\vspace{-5mm}
\end{figure*}

\subsubsection{Global vs. Local rewards}
Single-agent \gls{RL} environments have a single reward definition specifying the objective for the agent. Cooperative multi-agent settings can either have a single global reward or multiple local rewards, one for each agent~\cite{Wang20}. By contrast, global rewards only depend on the global state. They are easy to compute and specify but often lead to inefficient training because agents are never explicitly rewarded or penalized for their own actions~\cite{wol01}. Local rewards can reflect each agent's contribution to the cooperating group. This makes it easier to learn and improves convergence, especially in decentralized settings that do not have other explicit cooperation~\cite{Balch99}. However, it is usually challenging to design and implement a reward function that correctly rewards individual agents for their contributions~\cite{Hern2019}. Algorithms that aim to solve the problem of credit assignment described in Sec.~\ref{sec:algo-creditassignment} explicitly learn how global rewards can be disentangled into local rewards to solve this.

\section{Algorithms for MARL}
\label{sec:marl-algos}

We organize algorithms for \gls{MARL} into four categories based on their underlying paradigm, as illustrated in Fig.~\ref{fig:paradigms}.  
We order them from independent to fully connected agents.
The four paradigms described in the next paragraph are:
Fully decentralized algorithms, 
credit assignment, 
communicating agents, 
algorithms with centralized training and decentralized execution.
All algorithms surveyed in this work are summarized in Table~\ref{table:algorithms}. In addition to  paradigm and algorithm type, the table also lists available open-source implementations of these algorithms.

\begin{table*}
\caption{
We sort algorithms into the paradigms explained in Fig.\ref{fig:paradigms} according to their primary contribution. Some algorithms can be assigned to multiple paradigms. Additionally, we note if algorithms make use of an explicit communication mechanism between agents. The type is either value-based (V), policy-based (P), or actor-critic (AC).
}
\vspace{-1mm}
\label{table:algorithms}
\begin{center}
\begin{tabular}{p{8cm}rllp{4,5cm}}
\toprule
Name &  Year & Communication & Type & Framework \\
\midrule
\multicolumn{4}{l}{\textbf{Independent Agents}, Section~\ref{sec:algo-decentralized}} \\ 
IQL \cite{tan93} &  1993 &            No &    V &  EpyMARL\footnotemark[1], RLLib\footnotemark[2], MALib\footnotemark[4]\\
IA2C \cite{chu20} &  2020 &            No &    AC & EpyMARL\footnotemark[1], RLLib\footnotemark[2], MALib\footnotemark[4] \\
IPPO \cite{Witt2020} &  2020 &            No &    AC &EpyMARL\footnotemark[1], RLLib\footnotemark[2], MALib\footnotemark[4] \\
\midrule
\multicolumn{4}{l}{\textbf{Credit Assignment}, Section~\ref{sec:algo-creditassignment}} \\ 
COMA \cite{Foe2018} &  2018 &            No &   AC &  EpyMARL\footnotemark[1] \\
Shapley counterfactual credit assignment \cite{Li2021} &  2021 & No  &  Value  &        -/-      \\
VDN \cite{Sun18} &  2018 &            No &    V &             EpyMARL\footnotemark[1], Mava\footnotemark[3] \\
QMIX \cite{Rash18} &  2018 & No &V & EpyMARL\footnotemark[1], RLLib\footnotemark[2], Mava\footnotemark[3], MALib\footnotemark[4]\\
MAVEN \cite{Mah19} &  2019 &            No &  Hybrid &             Author's implementation\footnotemark[5] \\
Q-DPP \cite{Yang2020} &  2020 &            No &    V &  Author's implementation\footnotemark[6] \\
\midrule
\multicolumn{4}{l}{\textbf{Communication}, Section~\ref{sec:algo-communication}} \\ 
RIAL \cite{Foerster2016} &  2016 &           Yes &    V & Author's implementation\footnotemark[7] \\
DIAL \cite{Foerster2016} &  2016 & Yes &    V &   Mava\footnotemark[3], NMARL\footnotemark[8]  \\
CommNet \cite{Sukh2016} &  2016 &           Yes &    P &                     NMARL\footnotemark[8] \\
ATOC \cite{Jiang18} &  2018 &           Yes &   AC &                       -/- \\
IC3Net \cite{Singh2019} &  2019 &           Yes &  P & Author's implementation\footnotemark[9] \\
NeurComm \cite{chu2020} &  2020 &           Yes &  AC &  NMARL\footnotemark[8] \\
Mean-field MARL \cite{Yang2018} &  2018 &           No &  V/AC & Author's implementation\footnotemark[10] \\
Dec-Neural-AC \cite{Gu2021} & 2021 & No & AC & -/- \\
\midrule
\multicolumn{4}{l}{\textbf{Centralized Training, Decentralized Execution}, Section~\ref{sec:algo-cooperation}} \\ 
MADDPG \cite{Lowe2017} &  2017 & No & AC & EpyMARL\footnotemark[1], RLLib\footnotemark[2], MALib\footnotemark[4],  Mava\footnotemark[3] \\
MAPPO \cite{Yu2021} &  2021 &            No &   AC &           EpyMARL\footnotemark[1], Mava\footnotemark[3] \\
Recurrent policy optimization \cite{Kargar2021Macrpo} & 2021 & No & AC & Authors implementation\footnotemark[11]\\


\bottomrule
\end{tabular}
\end{center}
\footnotesize {
\begin{minipage}[t]{0.49\textwidth}
\footnotemark[1] \url{github.com/uoe-agents/epymarl}

\footnotemark[2] \url{docs.ray.io/en/master/rllib/index.html}

\footnotemark[3] \url{github.com/instadeepai/Mava}

\footnotemark[4] \url{github.com/sjtu-marl/malib}

\footnotemark[5] \url{github.com/AnujMahajanOxf/MAVEN}

\footnotemark[6] \url{github.com/QDPP-GitHub/QDPP}

\end{minipage}
\begin{minipage}[t]{0.49\textwidth}

\footnotemark[7] \url{github.com/iassael/learning-to-communicate}

\footnotemark[8] \url{github.com/cts198859/deeprl_network}

\footnotemark[9] \url{github.com/IC3Net/IC3Net}

\footnotemark[10] \url{github.com/mlii/mfrl}

\footnotemark[11] \url{github.com/kargarisaac/macrpo}

\end{minipage}
}
\vspace{-4mm}
\end{table*}

\textbf{Algorithm Frameworks.}
\red{There are many publicly available implementations of MARL algorithms, provided both through author implementations and larger frameworks.}
RLLib~\cite{rllib} is an open-source reinforcement learning library that aims at providing high-performance implementations for \gls{RL} algorithms. 
Many single-agent RL algorithms are implemented, which can be extended to the multi-agent setting. Moreover, some MARL-specific algorithms are available. EpyMARL is another framework comprising a wide range of algorithms. However, only discrete actions are available \cite{Papo2021}.
Mava is designed for scalable MARL training and execution. It allows for discrete  and continuous actions \cite{Pret2021}. 
MALib is specialized in scalable population-based MARL \cite{Zhou2021}. 
Networked \gls{MARL} (NMARL) is a repository of MARL algorithms for networked system control \cite{chu2020}.
Table~\ref{table:environments} lists the implementation framework, if available, as well.

\subsection{Decentralized: Learning independently} \label{sec:algo-decentralized}
The fully decentralized setting assumes that agents learn entirely independently. Usually, multiple agents learn individual single-agent \gls{RL} algorithms concurrently. To that end, each agent has an independent observation, policy, and algorithm - the only interaction exists through the environment.
This kind of algorithm is often used as a baseline and includes \gls{IQL}~\cite{tan93}, \gls{IA2C}~\cite{chu20} or \gls{IPPO}~\cite{Witt2020}.
Typically, independent agents perform worse than optimized cooperating agents in centralized settings. However, the performance benefits of more complicated approaches vary wildly between environments - in some cases, independent agents are on par with or can outperform cooperating agents~\cite{tan93, Witt2020}. 

While it is easy to adapt single-agent \gls{RL} algorithms to these settings, decentralized environments violate the stationary Markov property (i.e., the probability transitions $P$, which define the system dynamics, change). In particular, all other agents are considered part of the environment from an agent's perspective. Therefore, contrary to single-agent RL, convergence is not guaranteed \cite{Claus98}. However, decentralized algorithms do not need to deal with scalability issues \blue{which is a frequent challenge in autonomous mobility scenarios.}

\subsection{Credit assignment} \label{sec:algo-creditassignment}
Most \gls{MARL} environments do not explicitly disentangle reward functions for different agents and use global rewards instead. This is problematic during training~\cite{wol01}. Credit assignment methods aim to convert this global reward into an estimated, per-agent local reward. For example, if vehicles are trained to move autonomously, and a vehicle causes a crash, it is desirable to only punish the perpetrator. 

Credit assignment methods can be grouped into explicit and implicit credit assignment methods. Explicit methods estimate each agent's contributions against a certain baseline.
COMA~\cite{Foe2018} proposes a counterfactual baseline, where the global reward is compared to the reward received if the agent's action is replaced by a default action. This approach omits correlations between the agents, leading to inefficiency in complex settings. To circumvent this problem, Li et al.~\cite{Li2021} introduce Shapley counterfactual credit assignment that guides the training by estimating the individual credit for each agent using Shapley Values. Shapley Value is a concept from cooperative game theory quantifying the contribution of each player to the coalition \cite{shapley1953value}.

Implicit credit assignment aims at learning how to decompose the shared reward function into individual value functions \cite{Zhou20}. Sun et al.\ introduce a value decomposition network (VDN), which learns to decompose the shared value function into individual value functions \cite{Sun18}. However, it assumes additive individual functions. QMIX~\cite{Rash18} builds on VDN and removes this limitation by decomposing the global reward into an arbitrary non-linear combination of per-agent value functions. To make the optimization tractable, QMIX only constrains this combination to be monotonous with respect to the contribution of individual agents~\cite{Rash18}.

While this monotony constraint allows a decentralized execution, Mah et al.~\cite{Mah19} argue that it does not hold for arbitrary Q-functions and makes exploration inefficient. They introduce a hierarchical method, i.e., multi-agent variational exploration (MAVEN), that introduces a central latent space and diversifies exploration among agents~\cite{Mah19}. However, recent work~\cite{Yang2020} argues that this exploration is still inadequate and proposes a linear-time sampler that helps agents cover orthogonal directions in state space during training.

\subsection{Learning Communication} \label{sec:algo-communication}
To allow agents to cooperate, a large body of work introduces an explicit communication layer to the algorithms. This information exchange allows agents to widen their observation space and cooperate. We discuss algorithms that use explicit communication between agents.

Foerster et al.~\cite{Foerster2016} proposed two approaches for learning communication between independent agents. In Reinforced Inter-Agent Learning (RIAL), each agent has an additional network to generate a communication message for the other agents. For Differentiable Inter-Agent Learning (DIAL), the gradients of the other agents are fed through the communication channel into the communication network. This allows for end-to-end training across the agents. RIAL is only trainable within an agent. In contrast to discrete connections in RIAL and DIAL, CommNet~\cite{Sukh2016} learns a continuous communication channel jointly with the policy. Through this channel, the agents receive the summed transmission of the other agents. Then, the input of each hidden layer for each agent is the previous layer and the communication message. 

In recent work, communication has been viewed more ambiguously. Communication introduces costs, especially as the number of agents grows. Additionally, communication massively increases the amount of information available to each agent, making it harder to identify crucial information~\cite{Jiang18}. Recent work has thus focused on different mechanisms to decide if an agent should communicate at all.

Jiang and Lu~\cite{Singh2019} propose an attentional communication model (ATOC) that uses local information to decide if an agent should initiate communication. Agents form a dynamic group of local communicating agents that can coordinate with a shared communication medium. In a different line of work, the individualized controlled continuous communication model (IC3Net)~\cite{Singh2019} uses \gls{RNN} policies and allows agents to share the RNN's hidden state through a designated action. This is particularly useful in competitive scenarios as agents can block communication.

Another approach to restricting communication is networked \gls{MARL}. 
Contrary to attentional communication, these networks limit the agents' communication to their local neighbors. 
Chu et al.~\cite{chu2020} employ this concept and additionally introduce a communication protocol called NeurComm., It is similar to CommNet~\cite{Sukh2016} but instead of summing the received messages, NeurComm concatenates them. Policy fingerprints supplement communication to reduce non-stationarity \cite{chu2020}.


Representing each agent as a node in a network enables scaling to large agent populations. Yang et al.~\cite{Yang2018} propose employing this mean-field formulation to solve scalability issues. Accordingly, each agent is only affected by the mean effect of its local neighborhood. Compared to considering all agents, the complexity of interactions is reduced significantly for large agent populations. Decentralized neural actor-critic algorithm (Dec-Neural-AC)~\cite{Gu2021} is another mean-field approach aiming at scalability.



\subsection{Centralized Training with Decentralized Execution}
\label{sec:algo-cooperation}

Modern \gls{MARL} approaches often use centralization during training but create independent agents that run independently. This setting is known as \gls{CTDE}. \gls{CTDE} maps well to real-world use cases, where training can occur in simulation or with (potentially expensive) real-time communication, but agents should operate independently after training.

Usually, CTDE algorithms use an actor-critic architecture, which decomposes the policy and value estimates into explicit actor and critic networks. The critic is only used during training and to improve the policy's gradient updates. \gls{CTDE} actor-critic algorithms exploit this by training multiple agents together with a shared critic. Only the independent actors are needed during execution, allowing agents to operate without communication after training.

\Gls{MADDPG}~\cite{Lowe2017} is an example of \gls{CTDE} algorithms using per-agent actors with policy networks to map agent-specific observations to agent-specific actions. The critic approximates a centralized action-value function and receives the concatenated actions of all agents and additional global state information. This state information can include the concatenated observations of all agents. The agent's actions are included to make the environment more stationary~\cite{Lowe2017}.

Off-policy methods, such as \gls{MADDPG}, were commonly considered to be more sample-efficient than on-policy methods. However, Yu et al.~\cite{Yu2021} propose the on-policy algorithm \gls{MAPPO} that performs comparably to off-policy algorithms. \gls{MAPPO} suggests using global environment information as input to the critic instead of a concatenation of all observations, which scales with an increasing number of agents.

Recurrent Policy Optimization~\cite{Kargar2021Macrpo} is a recent method using a recurrent critic and a combinator that combines different agents' trajectories into one meta-trajectory. This meta-trajectory is used to train the critic and allows the approach to learning better from agents' interactions.

\section{MARL in Autonomous Mobility: Applications and Challenges}
\label{sec:applications}
Many real-world scenarios in autonomous mobility involve multiple agents. This presents an obvious potential for \gls{MARL} approaches to optimize how agents cooperate and interact. \blue{Autonomous mobility scenarios face similar key challenges. Large agent populations cause scalability and credit assignment issues. As mobility scenarios are usually highly dynamic, fast reaction times are required.} We give an overview of application domains of \gls{MARL} in autonomous mobility and discuss potential benefits, challenges, and recommendations for \gls{MARL}-based methods.
Table \ref{table:environments} lists different environments and their use-cases.
 
\begin{table*}
\caption{Simulation environments for general-purpose MARL research and four application fields summarized in Sec.~\ref{sec:applications}. Environments are characterized by either partial (P) or full (F) observability. Action spaces can either be discrete (D) or continuous (C). Slashes (A/B) denote that the environment can be configured to have different characteristics.}
\label{table:environments}
\begin{center}
\vspace{-2mm}
\begin{tabular}{llllp{8.5cm}}
\toprule
Environment &  Type & 
Obs. 
& 
Actions &                                             Description \\
\midrule
\textbf{General Purpose} \\
MAgent\footnotemark[1] \cite{Zheng18} &  Comp &  P &  D &     Multiple environments with a large number of agents\\
Multi-Agent Particle\footnotemark[1] \cite{mordatch2017emergence} &                         Coop / Comp &             P &     D / C &  Multiple environments focused on communication \\
\midrule
\textbf{Traffic Control} \\
CityFlow \cite{ZhangFLDZZ00JL19} & Coop / Comp  &           P / F &     -/- &                             City traffic simulator \\
SUMO \cite{SUMO2018} & Coop / Comp  &           P / F &      D / C &                                 Traffic simulation \\
\midrule
\textbf{Autonomous Vehicles}  \\
SMARTS\footnotemark[2] \cite{zhou2020smarts} & Coop &             P &       C &             Autonomous driving simulation platform \\
BARK\footnotemark[3] \cite{BernhardEHK20} &  Coop  &           P / F &     D / C &                       Autonomous driving scenarios  \\
MACAD\footnotemark[4] \cite{Pala2020} &                         Coop / Comp &             P &    D / C  &  Multi-Agent Connected Autonomous Driving built based on CARLA \\
highway-env\footnotemark[5] \cite{highway-env} & Coop  &           P / F &     D / C & Autonomous driving and tactical decision-making tasks \\
\midrule                
\textbf{Unmanned Aerial Vehicles} \\
Gym PyBullet Drones\footnotemark[6] \cite{panerati2021learning} & Coop / Comp &             P &       C &                                        Simulation environment for quadcopters with OpenAI gym API  \\
AirSim\footnotemark[7] \cite{shah2017airsim} & Coop / Comp & P & C & High-fidelity simulation environment for UAVs and autonomous vehicles \\
\midrule
\textbf{Resource Optimization} \\
                            MARO\footnotemark[7] \cite{MARO_MSRA} &  Coop  &             P &    D / C &        Resource optimization in industrial domains \\
                        Flatland\footnotemark[8] \cite{Mohan2020} &                               Coop &           P / F &       D &                      Vehicle re-scheduling problem             \\
\bottomrule
\end{tabular}

\raggedright
\centering
\footnotesize{
\vspace{0.2em}
\begin{minipage}[t]{.49\textwidth}
\footnotemark[1] \url{github.com/Farama-Foundation/PettingZoo}.

\footnotemark[2] \url{github.com/huawei-noah/SMARTS}

\footnotemark[3] \url{github.com/bark-simulator/bark-ml}

\footnotemark[4] \url{github.com/praveen-palanisamy/macad-gym}
\end{minipage}
\begin{minipage}[t]{.49\textwidth}
\footnotemark[5] \url{github.com/eleurent/highway-env}

\footnotemark[6] \url{github.com/utiasDSL/gym-pybullet-drones}

\footnotemark[7] \url{github.com/microsoft/maro}

\footnotemark[8] \url{flatland.aicrowd.com/intro.html}
\end{minipage}
}
\end{center}
\vspace{-6mm}
\end{table*}

\subsection{Traffic Control}
Traffic congestion in cities is problematic as it causes pollution, financial loss and increases the risk of accidents \cite{Wang21tlc}. Hence, controlling the traffic through traffic signals, line controls, or routing guidance has great potential to improve living conditions. Traffic involves many participants, including traffic lights, vehicles, and pedestrians. \gls{MARL} is an obvious fit for the traffic control problem as it does not rely on limiting heuristics. Data from various sources, e.g., road surveillance cameras, location-based mobile services, and vehicle tracking devices can be input into \gls{MARL} models~\cite{ZhangFLDZZ00JL19}.

Adaptive traffic signal control is a prominent approach to traffic control. A large number of agents, however, demand solutions to problems related to the scalability issue. 
As centralized and \gls{CTDE} approaches do not scale well for large agent populations, decentralized approaches are an evident choice. However, convergence is not guaranteed (Section \ref{sec:algo-decentralized}). To make the environment stationary from the agents' perspectives, Wang et al.~\cite{wang2021traffic} introduce graph attention networks to learn the dynamics of the neighbor intersection's influences. Instead of attention networks, Chu et al.~\cite{chu20} propose communication to the neighboring agents to improve the convergence of decentralized training. In addition, a spatial discount factor is used to weigh neighboring agents' observation and reward signals.


In essence, all approaches assume that neighboring agents are most relevant to controlling local traffic. However, training \gls{MARL} models needs a lot of data, making the availability of accurate simulators crucial. SUMO~\cite{SUMO2018} and CityFlow~\cite{ZhangFLDZZ00JL19} provide macroscopic city-scale simulations that can efficiently model many different agents and their interactions.





\subsection{Autonomous Vehicles}
A different application directly controls multiple autonomous vehicles through \gls{MARL}. Compared to independent single-agent RL \cite{schmidt2021trust}, this improves the ability of agents to cooperate and achieve their individual objectives.
In this domain, performance gains can come from coordinated driving (e.g., by reducing wind resistance, traffic jams, and optimizing road usage~\cite{chu2020}), the ability to extend the own perception with information from other agents (through vehicle-to-vehicle (V2V) / vehicle-to-X (V2X) communication~\cite{GeLL17}), or through efficient interactions in intersection navigation scenarios~\cite{wu2019dcl}.

A large variety of simulation and benchmark environments are available for research on autonomous vehicles.
The Highway-Env~\cite{highway-env} allows to control multiple vehicles and is particularly easy to set up and use. However, we found that performance can drop below acceptable levels for large-scale traffic simulations of many vehicles.
BARK~\cite{BernhardEHK20} is an open-source simulation framework for autonomous vehicle research focusing on multi-agent scenarios with interactions between agents. BARK simulates various traffic scenarios and has a two-way interface to control cars hosted in CARLA, a widely-used AV research simulation framework.
Similarly, SMARTS~\cite{zhou2020smarts} provides a simulation environment for diverse driving interactions, is compatible with the OpenAI gym interface, and offers multiple different observations.

This line of work presents unique challenges: Autonomous vehicles must act in dynamic and unpredictable environments like urban or highway traffic, and safety constraints must be satisfied at any time. This limits the exploration crucial for RL. Post-hoc extraction of safe policies trained in a simulator~\cite{schmidt2021trust} can be used to assure safety, but these methods have not yet been successfully applied to complex, multi-agent control settings. Moreover, most \gls{MARL} methods assume identical or compatible algorithms. This requires manufacturers to agree on a common architecture and communication standard to facilitate cross-brand cooperation of vehicles. 




\subsection{Unmanned Aerial Vehicles}
Swarms of \glspl{UAV} are another popular application field. UAVs are especially beneficial for communication tasks~\cite{cui2019application} and applications where human lives are endangered.
Here, a (potentially large) network of \glspl{UAV} needs to be controlled to fulfill tasks like forest fire surveillance \cite{CasbeerKBM06}, road traffic monitoring, \cite{ElloumiDEIS18} and air quality monitoring \cite{YangZBSH18}. 
Moreover, UAVs can be used to provide internet access in remote areas and enable wireless communication for applications like navigation and control \cite{MozaffariSBND19}.

Multiple works use \gls{MARL} to plan paths and assign targets for UAVs, in a problem known as Multiple Target Assignment and Path Planning (MUTAPP).
Here, MARL-based methods can outperform classical optimization techniques, like mixed-integer linear programming, because they can handle dynamic environments and operate in a decentralized way~\cite{QieSSXLW19}.
Qie et al.~\cite{QieSSXLW19} propose a solution based on MADDPG to jointly optimize target assignment and path planning. They propose a task abstraction layer that combines these tasks into a common reward structure. Their algorithm can effectively minimize the number of conflicts that arise in the MUTAPP problem.

In general, UAVs present a challenging problem for almost all algorithms. Typical environments for UAVs are highly dynamic~\cite{QieSSXLW19} and require constant collision avoidance (and thus quick reaction times) between individual agents. In addition, UAVs have strict real-time constraints for their operation, which limits complex central path planning and possible architectures for \gls{MARL} algorithms.~\cite{QieSSXLW19}.
Another major challenge is the energy supply and the limited range of UAVs.
Jung et al.~\cite{Jung21} apply a solution based on CommNet~\cite{Sukh2016} to coordinate the assignment of UAVs to charging towers and to share energy between towers, optimizing power draw from the electrical grid and minimizing operational costs.



Most research on UAVs uses custom, low-fidelity environments to simplify development~\cite{QieSSXLW19, Jung21}. However, multiple high-fidelity simulators are available that model UAV dynamics in more detail. AirSim~\cite{shah2017airsim} is a simulation platform based on the Unreal Engine with a special focus on realistic simulation and hardware-in-the-loop research. It supports detailed rendered camera observations. In addition to UAVs, AirSim is also able to simulate cars. Gym PyBullet Drones~\cite{panerati2021learning} presents a physical simulator for research on UAVs. In contrast to AirSim, Pybullet Drones features more realistic physical effects like the ground effect and downwash, and it is compatible with OpenAI gym and RLlib.

\subsection{Resource optimization}
In another line of work, \gls{MARL} is applied to resource optimization and scheduling for mobility scenarios. This includes scheduling for trains~\cite{Mohan2020} or ambulances, but also taxi repositioning~\cite{Liu21meta} and ride-sharing services~\cite{LiQJYWWWY19},
 bike repositioning, or container inventory management~\cite{MARO_MSRA}.
Li et al.~\cite{LiQJYWWWY19} propose to solve the assignment of ride requests to specific drivers with \gls{MARL}, which improves order response rates.


A popular simulation used in the train scheduling and routing domain is Flatland, which has been used in several NeurIPS competitions~\cite{Mohan2020}. Flatland provides a 2D-gridworld of train tracks. Agents (i.e., trains) have to find optimal policies that are safe, performant, and robust to unforeseen circumstances since trains can break down. Observations can be global or local, and Flatland also allows custom observation spaces.
The Multi-Agent Resource Optimization platform~\cite{MARO_MSRA} provides simulations, an RL toolkit, and utilities for distributed computing for this domain. It supports diverse scenarios across multiple application domains, including the aforementioned container inventory management and bike repositioning tasks, and includes visualizations of these simulation environments.

\glsresetall
\section{Conclusion}
\label{sec:conclusion}

In conclusion, we would like to note several challenges and open problems in \gls{MARL} research for mobility scenarios.
Chief among these is a lack of explicit safety and interpretability affordances in almost all current algorithms. This introduces real risk in autonomous mobility scenarios, especially when autonomous vehicles or UAVs are controlled.
Safe training and verification methods that certify policy safety~\cite{kontes2020high,schmidt2021trust,Rietsch2022} must be developed to deploy these methods in the real world~\cite{osinsky2020}.

A second challenge is integration with existing, manually controlled systems. This is evident in domains like traffic control or resource optimization, where human interaction and intervention (e.g., through emergency vehicles or manually prioritized resources) can come unexpectedly for agents that were only trained in a simulation. Effectively providing these interactions with human needs in simulation is an unsolved problem due to the low sample efficiency of RL.

Finally, most simulation environments assume perfect, high-bandwidth, latency-free communication. Despite recent advances in communication standards like 5G~\cite{GeLL17, Stahlke2022}, this is not a valid assumption in real-world scenarios. This makes more centralized solutions, like \gls{CTDE}, harder or impossible to train in real-world situations. Recent methods that limit communication could be a possible solution to this problem.

Overall, however, \gls{MARL} presents an exciting opportunity to learn solutions for complex problems involving multiple agents in an efficient, automated way.

\section*{Acknowledgements}
This work was supported by the Bavarian Ministry for Economic Affairs, Infrastructure, Transport and Technology through the Center for Analytics-Data-Applications (ADA-Center) within the framework of "BAYERN DIGITAL II". B.M.E. gratefully acknowledges support of the German Research Foundation (DFG) within the framework of the Heisenberg professorship program (Grant ES 434/8-1).

\bibliographystyle{IEEEtran}
\bibliography{references}

\begin{thebibliography}{10}
\providecommand{\url}[1]{#1}
\csname url@samestyle\endcsname
\providecommand{\newblock}{\relax}
\providecommand{\bibinfo}[2]{#2}
\providecommand{\BIBentrySTDinterwordspacing}{\spaceskip=0pt\relax}
\providecommand{\BIBentryALTinterwordstretchfactor}{4}
\providecommand{\BIBentryALTinterwordspacing}{\spaceskip=\fontdimen2\font plus
\BIBentryALTinterwordstretchfactor\fontdimen3\font minus
  \fontdimen4\font\relax}
\providecommand{\BIBforeignlanguage}[2]{{%
\expandafter\ifx\csname l@#1\endcsname\relax
\typeout{** WARNING: IEEEtran.bst: No hyphenation pattern has been}%
\typeout{** loaded for the language `#1'. Using the pattern for}%
\typeout{** the default language instead.}%
\else
\language=\csname l@#1\endcsname
\fi
#2}}
\providecommand{\BIBdecl}{\relax}
\BIBdecl

\bibitem{Oroo19}
A.~Oroojlooyjadid and D.~Hajinezhad, ``A review of cooperative multi-agent deep
  reinforcement learning,'' \emph{arXiv preprint 1908.03963}, 2019.

\bibitem{mordatch2017emergence}
I.~Mordatch and P.~Abbeel, ``Emergence of grounded compositional language in
  multi-agent populations,'' \emph{arXiv:1703.04908}, 2017.

\bibitem{sutton1998reinforcement}
R.~S. Sutton and A.~G. Barto, \emph{Reinforcement learning - an introduction},
  ser. Adaptive computation and machine learning.\hskip 1em plus 0.5em minus
  0.4em\relax {MIT} Press, 1998.

\bibitem{mnih2015human}
V.~Mnih, K.~Kavukcuoglu, D.~Silver, A.~A. Rusu, J.~Veness, M.~G. Bellemare,
  A.~Graves, M.~Riedmiller, A.~K. Fidjeland, G.~Ostrovski \emph{et~al.},
  ``Human-level control through deep reinforcement learning,'' \emph{Nature},
  vol. 518, no. 7540, pp. 529--533, 2015.

\bibitem{schulman2017proximal}
J.~Schulman, F.~Wolski, P.~Dhariwal, A.~Radford, and O.~Klimov, ``Proximal
  policy optimization algorithms,'' \emph{arXiv:1707.06347}, 2017.

\bibitem{Foerster2016}
J.~N. Foerster, Y.~M. Assael, N.~{De Freitas}, and S.~Whiteson, ``Learning to
  communicate with deep multi-agent reinforcement learning,'' in
  \emph{NeurIPS}, 2016, pp. 2145--2153.

\bibitem{Zhang2021}
K.~Zhang, Z.~Yang, and T.~Başar, ``Multi-agent reinforcement learning: A
  selective overview of theories and algorithms,'' in \emph{Studies in Systems,
  Decision and Control}, 2021, vol. 325, pp. 321--384.

\bibitem{Canese2021}
L.~Canese, G.~C. Cardarilli, L.~Di~Nunzio, R.~Fazzolari, D.~Giardino, M.~Re,
  and S.~Spanò, ``Multi-agent reinforcement learning: A review of challenges
  and applications,'' \emph{Appl. Sciences}, vol.~11, no.~11, 2021.

\bibitem{Litt2001}
M.~L. Littman, ``Value-function reinforcement learning in {M}arkov games,''
  \emph{Cognitive Systems Research}, vol.~2, no.~1, pp. 55--66, 2001.

\bibitem{SUMO2018}
P.~A. Lopez, M.~Behrisch, L.~Bieker-Walz, J.~Erdmann, Y.-P. Fl{\"o}tter{\"o}d,
  R.~Hilbrich, L.~L{\"u}cken, J.~Rummel, P.~Wagner, and E.~Wie{\ss}ner,
  ``Microscopic traffic simulation using sumo,'' in \emph{The 21st IEEE
  International Conference on Intelligent Transportation Systems}.\hskip 1em
  plus 0.5em minus 0.4em\relax IEEE, 2018.

\bibitem{ZhangFLDZZ00JL19}
H.~Zhang, S.~Feng, C.~Liu, Y.~Ding, Y.~Zhu, Z.~Zhou, W.~Zhang, Y.~Yu, H.~Jin,
  and Z.~Li, ``Cityflow: {A} multi-agent reinforcement learning environment for
  large scale city traffic scenario,'' in \emph{World Wide Web Conf.}, San
  Francisco, CA, 2019, pp. 3620--3624.

\bibitem{Wang20}
J.~Wang, Y.~Zhang, T.~K. Kim, and Y.~Gu, ``{Shapley Q-value:} a local reward
  approach to solve global reward games,'' in \emph{AAAI Conf.\ on Artificial
  Intelligence}, 2020, pp. 7285--7292.

\bibitem{wol01}
D.~H. Wolpert and K.~Tumer, ``Optimal payoff functions for members of
  collectives,'' \emph{Adv. in Compl. Sys.}, vol.~4, no. 2/3, pp. 265--279,
  2001.

\bibitem{Balch99}
T.~Balch, ``Reward and diversity in multirobot foraging,'' \emph{Work. Agents
  Learning About, From and With other Agents}, pp. 92--99, 1999.

\bibitem{Hern2019}
P.~Hernandez-Leal, B.~Kartal, and M.~E. Taylor, ``A survey and critique of
  multiagent deep reinforcement learning,'' \emph{Autonomous Agents and
  Multi-Agent Systems}, vol.~33, no.~6, pp. 750--797, nov 2019.

\bibitem{tan93}
M.~Tan, ``Multi-agent reinforcement learning: Independent vs. cooperative
  agents,'' \emph{Int. Conf. on Machine Learning}, pp. 330--337, 1993.

\bibitem{chu20}
T.~Chu, J.~Wang, L.~Codeca, and Z.~Li, ``Multi-agent deep reinforcement
  learning for large-scale traffic signal control,'' \emph{{IEEE} Trans.\
  Intell.\ Transportation Systems}, vol.~21, no.~3, pp. 1086--1095, Mar. 2020.

\bibitem{Witt2020}
C.~S. de~Witt, T.~Gupta, D.~Makoviichuk, V.~Makoviychuk, P.~H.~S. Torr, M.~Sun,
  and S.~Whiteson, ``Is independent learning all you need in the starcraft
  multi-agent challenge?'' \emph{arXiv preprint arXiv:2011.09533}, 2020.

\bibitem{Foe2018}
J.~N. Foerster, G.~Farquhar, T.~Afouras, N.~Nardelli, and S.~Whiteson,
  ``Counterfactual multi-agent policy gradients,'' in \emph{AAAI Conf.\ on
  Artificial Intelligence}, 2018, pp. 2974--2982.

\bibitem{Li2021}
J.~Li, K.~Kuang, B.~Wang, F.~Liu, L.~Chen, F.~Wu, and J.~Xiao, ``Shapley
  counterfactual credits for multi-agent reinforcement learning,'' in \emph{ACM
  Intl.\ Conf.\ Knowl. Disc. and Data Min.}, 2021, pp. 934--942.

\bibitem{Sun18}
P.~Sunehag, G.~Lever, A.~Gruslys, W.~M. Czarnecki, V.~Zambaldi, M.~Jaderberg,
  M.~Lanctot, N.~Sonnerat, J.~Z. Leibo, K.~Tuyls, and T.~Graepel,
  ``Value-decomposition networks for cooperative multi-agent learning based on
  team reward,'' in \emph{Intl. Jnt. Conf. Autonomous Agents and Multiagent
  Systems}, 2018, pp. 2085--2087.

\bibitem{Rash18}
J.~Foerster, S.~Whiteson, T.~Rashid, M.~Samvelyan, C.~{Schroeder De Witt}, and
  G.~Farquhar, ``{QMIX:} monotonic value function factorisation for deep
  multi-agent reinforcement learning,'' \emph{arXiv preprint
  arXiv:1803.11485v1}, 2018.

\bibitem{Mah19}
A.~Mahajan, T.~Rashid, M.~Samvelyan, and S.~Whiteson, ``{MAVEN:} multi-agent
  variational exploration,'' in \emph{NeurIPS}, 2019.

\bibitem{Yang2020}
Y.~Yang, Y.~Wen, J.~Wang, L.~Chen, K.~Shao, D.~Mguni, and W.~Zhang,
  ``Multi-agent determinantal q-learning,'' in \emph{Int.\ Conf.\ on Machine
  Learning}.\hskip 1em plus 0.5em minus 0.4em\relax PMLR, 2020, pp.
  10\,757--10\,766.

\bibitem{Sukh2016}
S.~Sukhbaatar, A.~Szlam, and R.~Fergus, ``Learning multiagent communication
  with backpropagation,'' in \emph{Advances in Neural Information Processing
  Systems}, 2016, pp. 2252--2260.

\bibitem{Jiang18}
J.~Jiang and Z.~Lu, ``Learning attentional communication for multi-agent
  cooperation,'' in \emph{NeurIPS}, 2018, pp. 7254--7264.

\bibitem{Singh2019}
A.~Singh, T.~Jain, and S.~Sukhbaatar, ``Learning when to communicate at scale
  in multiagent cooperative and competitive tasks,'' in \emph{Int.\ Conf.\ on
  Learning Representations}, 2019.

\bibitem{chu2020}
T.~Chu, S.~Chinchali, and S.~Katti, ``Multi-agent reinforcement learning for
  networked system control,'' \emph{arXiv preprint 2004.01339}, 2020.

\bibitem{Yang2018}
Y.~Yang, R.~Luo, M.~Li, M.~Zhou, W.~Zhang, and J.~Wang, ``Mean field
  multi-agent reinforcement learning,'' in \emph{Int.\ Conf.\ on Machine
  Learning}, vol.~12, 2018, pp. 5571--5580.

\bibitem{Gu2021}
H.~Gu, X.~Guo, X.~Wei, and R.~Xu, ``Mean-field multi-agent reinforcement
  learning: A decentralized network approach,'' \emph{arXiv preprint
  arXiv:2108.02731}, 2021.

\bibitem{Lowe2017}
R.~Lowe, Y.~Wu, A.~Tamar, J.~Harb, P.~Abbeel, and I.~Mordatch, ``Multi-agent
  actor-critic for mixed cooperative-competitive environments,'' in
  \emph{NeurIPS}, 2017, pp. 6380--6391.

\bibitem{Yu2021}
C.~Yu, A.~Velu, E.~Vinitsky, Y.~Wang, A.~Bayen, and Y.~Wu, ``The surprising
  effectiveness of ppo in cooperative, multi-agent games,'' \emph{arXiv
  preprint arXiv:2103.01955}, 2021.

\bibitem{Kargar2021Macrpo}
E.~Kargar and V.~Kyrki, ``{MACRPO:} multi-agent cooperative recurrent policy
  optimization,'' \emph{arXiv preprint 2109.00882}, 2021.

\bibitem{rllib}
E.~Liang, R.~Liaw, R.~Nishihara, P.~Moritz, R.~Fox, K.~Goldberg, J.~Gonzalez,
  M.~Jordan, and I.~Stoica, ``{RL}lib: Abstractions for distributed
  reinforcement learning,'' in \emph{Int.\ Conf.\ on Machine Learning}, 2018,
  pp. 3053--3062.

\bibitem{Papo2021}
G.~Papoudakis, F.~Christianos, L.~Sch{\"{a}}fer, and S.~V. Albrecht,
  ``Benchmarking multi-agent deep reinforcement learning algorithms in
  cooperative tasks,'' \emph{arXiv preprint arXiv:2006.07869}, 2020.

\bibitem{Pret2021}
A.~Pretorius, K.-A. Tessera, A.~P. Smit, C.~Formanek, S.~J. Grimbly, K.~Eloff,
  S.~Danisa, L.~Francis, J.~Shock, H.~Kamper, W.~Brink, H.~Engelbrecht,
  A.~Laterre, and K.~Beguir, ``Mava: a research framework for distributed
  multi-agent reinforcement learning,'' \emph{arXiv preprint arXiv:2107.01460},
  2021.

\bibitem{Zhou2021}
M.~Zhou, Z.~Wan, H.~Wang, M.~Wen, R.~Wu, Y.~Wen, Y.~Yang, W.~Zhang, and
  J.~Wang, ``{MALib:} a parallel framework for population-based multi-agent
  reinforcement learning,'' \emph{arXiv preprint arXiv:2106.07551}, 2021.

\bibitem{Claus98}
J.~Hao, D.~Huang, Y.~Cai, and H.~fung Leung, ``The dynamics of reinforcement
  social learning in networked cooperative multiagent systems,'' \emph{Eng.
  Applicat. of Artificial Intellig.}, vol.~58, pp. 111--122, 2017.

\bibitem{shapley1953value}
L.~Shapley, ``A value for n-person games,'' \emph{Ann. Math. Study28,
  Contributions to the Theory of Games}, pp. 307--317, 1953.

\bibitem{Zhou20}
M.~Zhou, Z.~Liu, P.~Sui, Y.~Li, and Y.~Y. Chung, ``Learning implicit credit
  assignment for cooperative multi-agent reinforcement learning,'' in
  \emph{Advances in Neural Information Processing Systems}, Dec. 2020.

\bibitem{Zheng18}
L.~Zheng, J.~Yang, H.~Cai, W.~Zhang, J.~Wang, and Y.~Yu, ``{MAgent:} a
  many-agent reinforcement learning platform for artificial collective
  intelligence,'' in \emph{AAAI Conf. Artificial Intellig.}, 2018, pp.
  8222--8223.

\bibitem{zhou2020smarts}
M.~Zhou, J.~Luo, J.~Villella, Y.~Yang, D.~Rusu, J.~Miao, W.~Zhang, M.~Alban,
  I.~Fadakar, Z.~Chen, A.~C. Huang, Y.~Wen, K.~Hassanzadeh, D.~Graves, D.~Chen,
  Z.~Zhu, N.~Nguyen, M.~Elsayed, K.~Shao, S.~Ahilan, B.~Zhang, J.~Wu, Z.~Fu,
  K.~Rezaee, P.~Yadmellat, M.~Rohani, N.~P. Nieves, Y.~Ni, S.~Banijamali, A.~C.
  Rivers, Z.~Tian, D.~Palenicek, H.~bou Ammar, H.~Zhang, W.~Liu, J.~Hao, and
  J.~Wang, ``{SMARTS:} scalable multi-agent reinforcement learning training
  school for autonomous driving,'' in \emph{Conf.\ on Robot Learning}, Nov.
  2020.

\bibitem{BernhardEHK20}
J.~Bernhard, K.~Esterle, P.~Hart, and T.~Kessler, ``{BARK:} open behavior
  benchmarking in multi-agent environments,'' in \emph{{IEEE/RSJ} Int.\ Conf.\
  Intelligent Robots and Systems}, Las Vegas, NV, 2020, pp. 6201--6208.

\bibitem{Pala2020}
P.~Palanisamy, ``Multi-agent connected autonomous driving using deep
  reinforcement learning,'' in \emph{Int.\ Joint Conf.\ on Neural Networks},
  2020.

\bibitem{highway-env}
E.~Leurent, ``An environment for autonomous driving decision-making,''
  \url{https://github.com/eleurent/highway-env}, 2018.

\bibitem{panerati2021learning}
J.~Panerati, H.~Zheng, S.~Zhou, J.~Xu, A.~Prorok, and A.~P. Schoellig,
  ``Learning to fly---a gym environment with pybullet physics for reinforcement
  learning of multi-agent quadcopter control,'' in \emph{IEEE/RSJ Int.\ Conf.\
  on Intelligent Robots and Systems (IROS)}, 2021.

\bibitem{shah2017airsim}
S.~Shah, D.~Dey, C.~Lovett, and A.~Kapoor, ``Airsim: High-fidelity visual and
  physical simulation for autonomous vehicles,'' in \emph{Intl.\ Conf.\ Field
  and Serv.\ Robot.}, Zurich, Switzerland, 2017, pp. 621--635.

\bibitem{MARO_MSRA}
X.~Li, J.~Zhang, J.~Bian, Y.~Tong, and T.-Y. Liu, ``A cooperative multi-agent
  reinforcement learning framework for resource balancing in complex logistics
  network,'' in \emph{Int.\ Joint Conf.\ on Autonomous Agents and Multiagent
  Systems}, Mar. 2019.

\bibitem{Mohan2020}
S.~Mohanty, E.~Nygren, F.~Laurent, M.~Schneider, C.~Scheller, N.~Bhattacharya,
  J.~Watson, A.~Egli, C.~Eichenberger, C.~Baumberger, G.~Vienken, I.~Sturm,
  G.~Sartoretti, and G.~Spigler, ``{Flatland-RL:} multi-agent reinforcement
  learning on trains,'' \emph{arXiv preprint arXiv:2012.05893}, 2020.

\bibitem{Wang21tlc}
Z.~Wang, H.~Zhu, M.~He, Y.~Zhou, X.~Luo, and N.~Zhang, ``Gan and multi-agent
  {DRL} based decentralized traffic light signal control,'' \emph{{IEEE}
  Trans.\ Veh.\ Technol.}, 2021.

\bibitem{wang2021traffic}
M.~Wang, L.~Wu, J.~Li, and L.~He, ``Traffic signal control with reinforcement
  learning based on region-aware cooperative strategy,'' \emph{IEEE
  Transactions on Intelligent Transportation Systems}, 2021.

\bibitem{schmidt2021trust}
L.~M. Schmidt, G.~Kontes, A.~Plinge, and C.~Mutschler, ``Can you trust your
  autonomous car? interpretable and verifiably safe reinforcement learning,''
  in \emph{{IEEE} Intelligent Vehicles Symp.}, Nagoya, Japan, Jul. 2021, pp.
  171--178.

\bibitem{GeLL17}
X.~Ge, Z.~Li, and S.~Li, ``{5G} software defined vehicular networks,''
  \emph{arXiv preprint arXiv:1702.03675}, 2017.

\bibitem{wu2019dcl}
Y.~Wu, H.~Chen, and F.~Zhu, ``Dcl-aim: Decentralized coordination learning of
  autonomous intersection management for connected and automated vehicles,''
  \emph{Transportation Research Part C: Emerging Technologies}, vol. 103, pp.
  246--260, 2019.

\bibitem{cui2019application}
J.~Cui, Y.~Liu, and A.~Nallanathan, ``The application of multi-agent
  reinforcement learning in {UAV} networks,'' in \emph{{IEEE} Int.\ Conf.\ on
  Communications Workshops}.\hskip 1em plus 0.5em minus 0.4em\relax Shanghai,
  China: {IEEE}, May 2019.

\bibitem{CasbeerKBM06}
D.~W. Casbeer, D.~B. Kingston, R.~W. Beard, and T.~W. McLain, ``Cooperative
  forest fire surveillance using a team of small unmanned air vehicles,''
  \emph{Int. J. Syst. Sci.}, vol.~37, no.~6, pp. 351--360, 2006.

\bibitem{ElloumiDEIS18}
M.~Elloumi, R.~Dhaou, B.~Escrig, H.~Idoudi, and L.~A. Sa{\"{\i}}dane,
  ``Monitoring road traffic with a uav-based system,'' in \emph{{IEEE} Wireless
  Communications and Networking Conf.}, 2018.

\bibitem{YangZBSH18}
Y.~Yang, Z.~Zheng, K.~Bian, L.~Song, and Z.~Han, ``Real-time profiling of
  fine-grained air quality index distribution using {UAV} sensing,''
  \emph{{IEEE} Internet Things J.}, vol.~5, no.~1, pp. 186--198, 2018.

\bibitem{MozaffariSBND19}
M.~Mozaffari, W.~Saad, M.~Bennis, Y.~Nam, and M.~Debbah, ``A tutorial on uavs
  for wireless networks: Applications, challenges, and open problems,''
  \emph{{IEEE} Commun. Surv. Tutorials}, vol.~21, no.~3, pp. 2334--2360, 2019.

\bibitem{QieSSXLW19}
H.~Qie, D.~Shi, T.~Shen, X.~Xu, Y.~Li, and L.~Wang, ``Joint optimization of
  multi-uav target assignment and path planning based on multi-agent
  reinforcement learning,'' \emph{{IEEE} Access}, vol.~7, pp.
  146\,264--146\,272, 2019.

\bibitem{Jung21}
S.~Jung, W.~J. Yun, J.~Kim, J.-H. Kim, and F.~Falcone, ``Coordinated
  multi-agent deep reinforcement learning for energy-aware {UAV}-based big-data
  platforms,'' \emph{mdpi.com}, 2021.

\bibitem{Liu21meta}
C.~Liu, C.~X. Chen, and C.~Chen, ``{META:} a city-wide taxi repositioning
  framework based on multi-agent reinforcement learning,'' \emph{{IEEE} Trans.\
  on Intelligent Transportation Systems}, 2021.

\bibitem{LiQJYWWWY19}
M.~Li, Z.~T. Qin, Y.~Jiao, Y.~Yang, J.~Wang, C.~Wang, G.~Wu, and J.~Ye,
  ``Efficient ridesharing order dispatching with mean field multi-agent
  reinforcement learning,'' in \emph{World Wide Web Conference}, San Francisco,
  CA, 2019, pp. 983--994.

\bibitem{kontes2020high}
G.~Kontes, D.~Scherer, T.~Nisslbeck, J.~Fischer, and C.~Mutschler, ``High-speed
  collision avoidance using deep reinforcement learning and domain
  randomization for autonomous vehicles,'' in \emph{{IEEE} Int.\ Conf.\
  Intell.\ Transportation Systems}, 2020.

\bibitem{Rietsch2022}
S.~Rietsch, S.-Y. Huang, G.~Kontes, A.~Plinge, and C.~Mutschler, ``Driver dojo:
  A benchmark for generalizable reinforcement learning for autonomous
  driving,'' \emph{arXiv preprint arXiv:2207.11432}, 2022.

\bibitem{osinsky2020}
B.~Osinski, A.~Jakubowski, P.~Ziecina, P.~Milos, S.~Galias, C.~Homoceanu, and
  H.~Michalewski, ``Simulation-based reinforcement learning for real-world
  autonomous driving,'' in \emph{{IEEE} Int.\ Conf.\ Robotics and Automation},
  May 2020, pp. 6411--6418.

\bibitem{Stahlke2022}
M.~Stahlke, T.~Feigl, M.~H. Castaneda~Garcia, R.~S. Gallacher, J.~Seitz, and
  C.~Mutschler, ``Transfer learning to adapt {5G} fingerprint-based
  localization across environments,'' in \emph{{IEEE} Veh.\ Tech.\ Conf.},
  2022.

\end{thebibliography}

\end{document}